\def\year{2021}\relax
\documentclass[letterpaper]{article} 
\usepackage{aaai21}  
\usepackage{times}  
\usepackage{helvet} 
\usepackage{courier}  
\usepackage[hyphens]{url}  
\usepackage{graphicx} 
\usepackage{booktabs}
\usepackage{listings}
\usepackage{xcolor}
\usepackage{todonotes}
\usepackage{cleveref}
\usepackage{natbib}  
\usepackage{caption} 
\definecolor{lightBlue}{RGB}{0, 153, 255}
\definecolor{lightRed}{RGB}{204, 0, 255}

\colorlet{punct}{red!60!black}
\definecolor{background}{HTML}{EEEEEE}
\definecolor{delim}{RGB}{20,105,176}
\colorlet{numb}{magenta!60!black}

\lstdefinelanguage{json}{
    basicstyle=\normalfont\ttfamily\fontsize{8}{9}\selectfont,
    numbers=left,
    numberstyle=\scriptsize,
    stepnumber=1,
    numbersep=8pt,
    showstringspaces=false,
    breaklines=true,
    frame=lines,
    backgroundcolor=\color{background},
    literate=
     *{0}{{{\color{numb}0}}}{1}
      {1}{{{\color{numb}1}}}{1}
      {2}{{{\color{numb}2}}}{1}
      {3}{{{\color{numb}3}}}{1}
      {4}{{{\color{numb}4}}}{1}
      {5}{{{\color{numb}5}}}{1}
      {6}{{{\color{numb}6}}}{1}
      {7}{{{\color{numb}7}}}{1}
      {8}{{{\color{numb}8}}}{1}
      {9}{{{\color{numb}9}}}{1}
      {:}{{{\color{punct}{:}}}}{1}
      {,}{{{\color{punct}{,}}}}{1}
      {\{}{{{\color{delim}{\{}}}}{1}
      {\}}{{{\color{delim}{\}}}}}{1}
      {[}{{{\color{delim}{[}}}}{1}
      {]}{{{\color{delim}{]}}}}{1},
}

\title{Exploring and Analyzing Machine Commonsense Benchmarks}
\author{
    Henrique Santos,
    Minor Gordon,
    Zhicheng Liang,
    Gretchen Forbush,
    Deborah L. McGuinness\\
}
\affiliations{
    Rensselaer Polytechnic Institute\\
    110 8th St, Troy, New York 12180\\
    \{oliveh,gordom6,liangz4,forbug\}@rpi.edu, dlm@cs.rpi.edu
}

\begin{document}

\maketitle

\begin{abstract}
Commonsense question-answering (QA) tasks, in the form of benchmarks, are constantly being introduced for challenging and comparing commonsense QA systems. The benchmarks provide question sets that systems’ developers can use to train and test new models before submitting their implementations to official leaderboards. Although these tasks are created to evaluate systems in identified dimensions (e.g. topic, reasoning type), this metadata is limited and largely presented in an unstructured format or completely not present. Because machine common sense is a fast-paced field, the problem of fully assessing current benchmarks and systems with regards to these evaluation dimensions is aggravated. We argue that the lack of a common vocabulary for aligning these approaches' metadata limits researchers in their efforts to understand systems' deficiencies and in making effective choices for future tasks. In this paper, we first discuss this MCS ecosystem in terms of its elements and their metadata. Then, we present how we are supporting the assessment of approaches by initially focusing on commonsense benchmarks. We describe our initial MCS Benchmark Ontology, an extensible common vocabulary that formalizes benchmark metadata, and showcase how it is supporting the development of a Benchmark tool that enables benchmark exploration and analysis.
\end{abstract}

\section{Introduction}

Machine commonsense (MCS) benchmarks have arisen as a way to challenge AI 
question answering systems by presenting a set of natural language questions that require these 
systems to solve tasks that involve what some perceive as commonsense knowledge. The MCS community is constantly introducing diversified tasks and allowing question-answering (QA) systems' developers to submit their systems to official leaderboards. These leaderboards have emerged to act as hubs for hosting benchmarks and supporting infrastructure that accepts submissions of systems that then get scored against these tasks. This MCS ecosystem consists of benchmarks and tasks, QA systems and models, structured commonsense knowledge, leaderboards, scientific publications, and the people and institutions behind these efforts. Although surveys exist~\cite{storks_recent_2020}, they quickly become outdated due to the field's quick pace.


The DARPA's Machine Common Sense Program\footnote{\url{https://www.darpa.mil/program/machine-common-sense}} is an ongoing effort that has been exploring the boundaries of the commonsense question-answering state-of-the-art, with the ultimate goal of creating a commonsense computational service that can solve diverse challenges. In support of this, the program is promoting the creation of new, improved, or specialized commonsense tasks from within the project as well as adopting challenging benchmarks from outside the project. In addition, it is supporting the augmentation and incorporation of structured commonsense knowledge in QA systems. Under the program, commonsense approaches are being categorized in terms of \textit{evaluation dimensions} that span across tasks and systems. They aim to capture relevant aspects from these elements, allowing approaches to be classified into common categories, supporting insights into the available tasks, and informing stakeholders about the commonsense aspects they explore. In addition, they support the program in making decisions for future tasks. The currently identified dimensions include \textit{question type} (e.g, multiple-choice, open-ended), \textit{topic} (e.g. social, events), and \textit{reasoning type} (e.g. temporal, spatial).

In this context, it is currently a challenge to efficiently perform analysis or extract insights into the dimensions of these composing elements that can support potential use-cases from the community. As an example, a principal investigator could ask, \textit{``what are the tasks that challenge systems in temporal reasoning?''} As benchmarks currently convey limited metadata, even more challenging is how to scale these dimensions to assess not only tasks as a whole, but as well as to assess each question. Benchmarks can contain questions that are varied, each of which focusing on specific dimensions.

In this paper, we demonstrate how we are tackling the problem of assessing MCS approaches by initially focusing on commonsense benchmarks. We introduce our MCS Benchmark ontology and describe how it is being applied in support of a Benchmark tool to enable the exploration and analysis of multiple commonsense tasks. The ontology specifies concepts that describe the datasets, making use of the Schema.org simplified taxonomy. Refined classes allow the description of several constructs that compose benchmarks. The ontology acts as a formalization of a subset of the currently identified evaluation dimensions within DARPA's MCS program.

\section{MCS Benchmark ontology}

The MCS Benchmark ontology aims to formalize the diverse benchmark metadata in a common vocabulary. In support of the ontology development, we have surveyed several state-of-the-art benchmarks to understand their constructs, and a summary is shown on  \Cref{tbl:benchmarks}. They were selected based on DARPA's adoption of these tasks for Year 1 of the MCS program.

\begin{table}[h!]
\resizebox{\columnwidth}{!}{
\begin{tabular}{@{}lll@{}}
\toprule
\textbf{Benchmark} & \textbf{Constructs}                                                                                         & \textbf{Question type}                                                     \\ \midrule
\begin{tabular}[c]{@{}l@{}}aNLI\\\cite{bhagavatula_abductive_2019}\end{tabular}               & \begin{tabular}[c]{@{}l@{}}- Observations\\ - Hypothesis\end{tabular}                                       & - Multiple choice                                                        \\ \midrule
\begin{tabular}[c]{@{}l@{}}CommonsenseQA\\\cite{talmor_commonsenseqa_2019}\end{tabular}      & \begin{tabular}[c]{@{}l@{}}- Questions\\ - Answer choices\end{tabular}                                      & - Multiple choice                                                        \\ \midrule
\begin{tabular}[c]{@{}l@{}}CosmosQA\\\cite{huang_cosmos_2019}\end{tabular}          & \begin{tabular}[c]{@{}l@{}}- Context\\ - Questions\\ - Answer choices\end{tabular}                          & - Multiple choice                                                        \\ \midrule
CycIC\footnote{\url{https://leaderboard.allenai.org/cycic/submissions/about}}              & \begin{tabular}[c]{@{}l@{}}- Questions\\ - Answer choices\\ - Fill in the blank\\ - Categories\end{tabular} & \begin{tabular}[c]{@{}l@{}}- Multiple choice\\ - True/false\end{tabular} \\ \midrule
\begin{tabular}[c]{@{}l@{}}HellaSwag\\\cite{zellers_hellaswag_2019}\end{tabular}          & \begin{tabular}[c]{@{}l@{}}- Context\\ - Ending choices\end{tabular}                                        & - Multiple choice                                                        \\ \midrule
\begin{tabular}[c]{@{}l@{}}Physical IQa\\\cite{bisk_piqa_2020}\end{tabular}       & \begin{tabular}[c]{@{}l@{}}- Goals\\ - Solution choices\end{tabular}                                        & - Multiple choice                                                        \\ \midrule
\begin{tabular}[c]{@{}l@{}}Social IQa\\\cite{sap_social_2019}\end{tabular}         & \begin{tabular}[c]{@{}l@{}}- Context\\ - Questions\\ - Answer choices\end{tabular}                          & - Multiple choice                                                        \\ \bottomrule
\end{tabular}
}
\caption{Benchmarks and their identified constructs.}
\label{tbl:benchmarks}
\end{table}

While all of the benchmarks provide ``multiple-choice'' questions (with CycIC providing a binary kind of multiple-choice: true/false), the benchmarks diverge on the kinds of constructs they include. CommonsenseQA provides a standard question/answer format. CosmosQA and Social IQa include the context on top of that. HellaSwag also provides context, but it requires systems to choose the more appropriate ending for the context, instead of answering a question. aNLI provides observations (usually a scene or a setting) and asks systems to explain the reasons for that observation to happen in the form of a hypothesis. Physical IQa provides goals (or objectives), alongside possible solutions to achieve them. CycIC is an interesting case as it provides an increased level of metadata for each sample. These include categories, which contain information about what kind of reasoning is needed, and/or question classification into common types.

\begin{figure}
  \includegraphics[width=\linewidth]{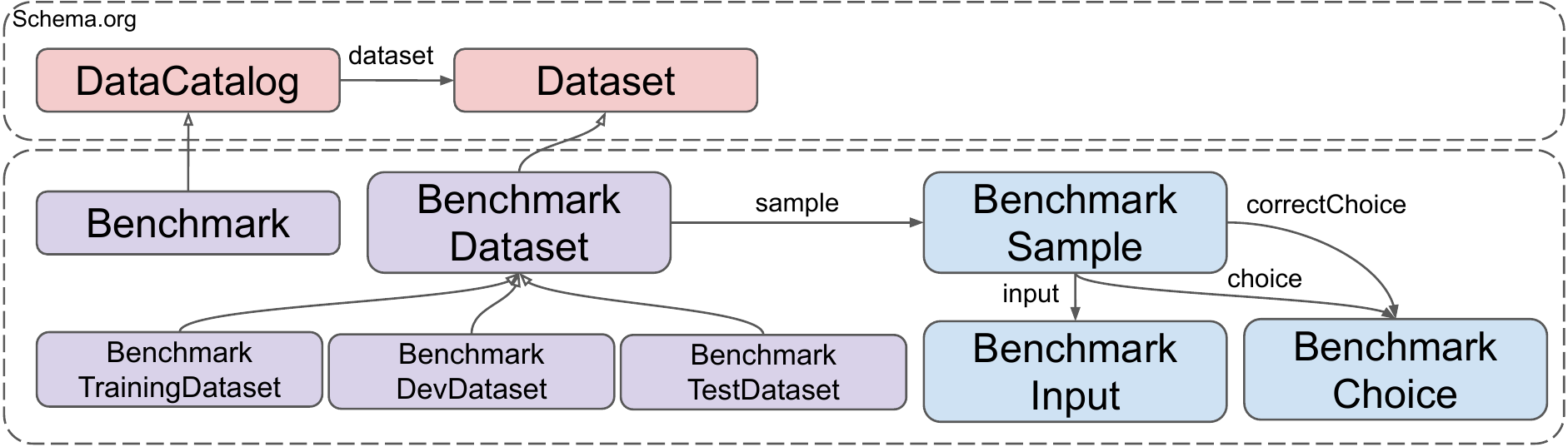}
  \centering
  \caption{Part of the MCS Benchmark Ontology}
  \label{fig:ontology}
\end{figure}

The MCS Benchmark ontology provides a common modeling for these diverse constructs, as seen in Figure~\ref{fig:ontology}. Each entry in a benchmark dataset is defined as a \texttt{BenchmarkSample}. Each sample is then composed of one or more instances of \texttt{BenchmarkInput}. Input is considered anything that a benchmark provides that systems can use. These are the constructs identified in Table~\ref{tbl:benchmarks}. Each sample also contains the possible choices, represented by the \texttt{BenchmarkChoice} class. The correct choice, which is used to train/verify proposed models, is linked to the sample by the \texttt{correctChoice} property.

\begin{lstlisting}[language=json,firstnumber=1,caption={Benchmark sample in JSON-LD},label=lst-sample]
{
  "@context": "https://.../context.jsonld",
  "@id": "SocialIQa-37",
  "@type": "BenchmarkSample",
  "includedInDataset": "SocialIQa/train",
  "input": [
    {
      "@id": "SocialIQa-37-input-0",
      "@type": "BenchmarkContext",
      "text": "Skylar returned early in the evening after a night and day of partying."
    },
    {
      "@id": "SocialIQa-37-input-1",
      "@type": "BenchmarkQuestion",
      "text": "How would you describe Skylar?"
    }
  ],
  "choice": [
    {
      "@id": "SocialIQa-37-choice-1",
      "@type": "BenchmarkAnswer",
      "text": "a party girl"
    },
    {
      "@id": "SocialIQa-37-choice-2",
      "@type": "BenchmarkAnswer",
      "text": "very shy"
    },
    {
      "@id": "SocialIQa-37-choice-3",
      "@type": "BenchmarkAnswer",
      "text": "exhausted"
    }
  ],
  "correctChoice": {
    "@id": "SocialIQa-37-choice-1"
  }
}
\end{lstlisting}

In \Cref{lst-sample}, we show an entry from the Social IQa benchmark represented using the MCS Benchmark ontology, in JSON-LD. The sample is composed of a list of inputs (a context and a question) and a list of choices (possible answers for the question). To assert the sample as part of either the training, development, or test dataset, the ontology defines a set of classes that are used to represent the benchmark datasets. The \texttt{includedInDataset} property links samples to instances of \texttt{BenchmarkTrainDataset}, \texttt{BenchmarkDevDataset}, and \texttt{BenchmarkTestDataset}.

\section{Evaluation: Supporting exploration and analysis of Benchmarks}

The MCS Benchmark ontology is used in support of a prototype Benchmark tool that provides several features for interacting with benchmarks from multiple sources. The Benchmark tool allows users to explore and analyze benchmarks by leveraging the common modeling provided by the ontology. To enable this, we have implemented a converter that can receive the datasets that compose benchmarks as input and, using the terminology in the ontology, it outputs them in the common JSON-LD format, as seen in \Cref{lst-sample}. To further simplify the JSON-LD serialization, we have encapsulated many linked data constructs (including namespaces and property types) in a JSON-LD context file (i.e. \texttt{@context} key). This allows us to suppress verbose information in the JSON, while keeping the correct and concise expression of the model.

\begin{figure}
  \includegraphics[width=\linewidth]{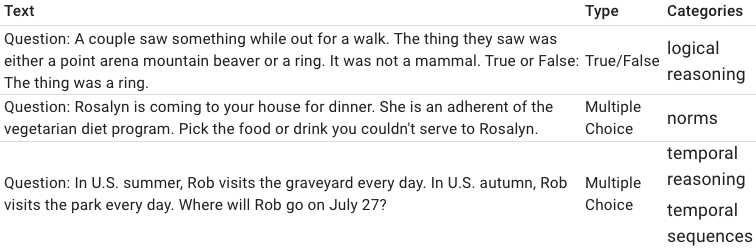}
  \centering
  \caption{Benchmark tool displaying questions and associated metadata}
  \label{fig:benchmark-tool}
\end{figure}

Figure~\ref{fig:benchmark-tool} contains a screen of the tool displaying some of the CycIC questions alongside the available metadata. We provide a sample Benchmark tool usage of the ontology in \Cref{lst-query01}. The SPARQL query retrieves training samples of a specific benchmark by constraining the dataset to be of type \texttt{BenchmarkTrainDataset}. For each sample, their input texts and types are retrieved.

\lstset{language=SQL,morekeywords={PREFIX,rdf,rdfs,mcs,schema}}
\begin{lstlisting}[
    firstnumber=1,caption={Querying training samples of a benchmark},label=lst-query01,
    basicstyle=\normalfont\ttfamily\fontsize{7}{8}\selectfont,
    numbers=left,
    numberstyle=\scriptsize,
    stepnumber=1,
    numbersep=8pt,
    showstringspaces=false,
    breaklines=true,
    frame=lines,
    backgroundcolor=\color{background}
]
SELECT ?sample ?input ?inputType WHERE {
 <task_uri> schema:dataset ?train .
 ?train rdf:type mcs:BenchmarkTrainDataset .
 ?train mcs:sample ?sample .
 ?sample mcs:input/schema:text ?input .
 ?sample mcs:input/rdf:type/rdfs:label ?inputType .
}
\end{lstlisting}

\Cref{lst-query02} shows a SPARQL query that retrieves samples containing logical reasoning questions across different benchmarks. The ontology represents each of the identified input types, therefore it enables querying by a specific type, in this case, \texttt{BenchmarkQuestion}.

\lstset{language=SQL,morekeywords={PREFIX,rdf,rdfs,mcs,schema}}
\begin{lstlisting}[
    firstnumber=1,caption={Querying question samples across benchmarks},label=lst-query02,
    basicstyle=\normalfont\ttfamily\fontsize{7}{8}\selectfont,
    numbers=left,
    numberstyle=\scriptsize,
    stepnumber=1,
    numbersep=8pt,
    showstringspaces=false,
    breaklines=true,
    frame=lines,
    backgroundcolor=\color{background}
]
SELECT ?sample ?question WHERE {
 ?sample rdf:type mcs:BenchmarkSample .
 ?sample mcs:input/rdf:type mcs:LogicalReasoning .
 ?sample mcs:input ?input .
 ?input rdf:type mcs:BenchmarkQuestion .
 ?input schema:text ?question .
}
\end{lstlisting}

\section{Related work}

Throughout the past decade, the challenge for Question Answering over Linked Data\footnote{\url{http://qald.aksw.org}} (QALD) has been used as a way of promoting the development of question-answering systems capable of solving benchmarks, with an increased difficulty based on the growing availability of linked data on the web. In its latest edition, participants were required to integrate their entrant systems with the GERBIL QA~\cite{usbeck_benchmarking_2019} benchmark platform. GERBIL QA serves as an integration service for QA systems, supporting the fair comparison of systems through the use of unified metrics and integrated benchmark datasets. GERBIL QA represents question sets in a JSON-based format that serves as an interface that characterizes each question, including data types, entities, and keywords. This format is highly tailored to question answering over linked data, as it assists systems in building responses that comply with the expected format.

AI Collaboratory~\cite{martinez-plumed_tracking_2020} has the objective of being a platform for the analysis and comparison of AI systems. Its scope goes beyond question answering, allowing submissions of systems that solve diverse and specialized AI tasks (e.g. link prediction, speech recognition, and more). Tasks are represented in an Entity-Relationship model without an in-depth formalization of tasks' metadata. AI2 Leaderboard,\footnote{\url{https://leaderboard.allenai.org}} maintained by the Allen Institute for AI, hosts many commonsense benchmarks and accepts submissions of systems. In this approach, benchmarks are stored as originally released alongside documentation (in natural language) that includes a description of the tasks, and the format of the datasets.

The AIDA Dashboard~\cite{angioni_aida_2020} is an implementation that assists editors of scientific publishers in assessing conferences in Computer Science, with regards to some dimensions, including citations, topic, and similar conferences. It uses the Computer Science Ontology (CSO) to annotate research papers with a common vocabulary and leverages this annotation to provide visualizations that extract insights based on these dimensions. Although it focuses on a different domain, this work is closely related to ours as it relies on a metadata formalization as an ontology that aligns dimensions in support of the creation of dashboards.

To the best of our knowledge, none of the previous attempted to formalize and integrate metadata across the variety of commonsense tasks. Our Benchmark tool, supported by the MCS Benchmark ontology, aims to bridge this gap in machine commonsense, where the lack of metadata formalization is constraining the ability of further assessing tasks in terms of identified dimensions.

\section{Conclusion}

In a fast-paced field, such as machine common sense, we argue that it is essential to promote the formalization of metadata to support the development and analysis of evaluation metrics. Further, a common representation metadata schema can support sharing and communication of results that can be compared and contrasted more easily, and can thus help the MCS community understand more about what the benchmarks well suited to test and how different approaches and methods may compare in different contexts. We presented our evolving MCS Benchmark ontology that is aimed to support our Benchmark exploration and analysis tool. Existing platforms for benchmarks are largely task-specific and limited in terms of metadata formalization. In commonsense reasoning, where the ultimate goal is to have a commonsense service that can actively learn from new and specialized tasks, there is a need to support the incorporation of these diverse benchmarks, while allowing implementations to access them in a standardized way. We believe a centralized platform, where the MCS community can obtain trends and comparisons across tasks, will greatly support this objective.

We are expanding our review of commonsense benchmarks, including open-ended benchmarks (benchmarks that are not multiple-choice, requiring systems to elaborate their answer in the form of a natural language sentence, e.g. CommonGen~\cite{lin_commongen_2020}), and we are expanding the ontology to support these. In parallel, we started to work towards the formalization of the evaluation dimensions that are related to QA systems, with a focus on \textit{interpretability}. To this extent, we are analyzing QA systems and identifying information that can be extracted during their execution and represented in the ontology, as a way of assessing a system. The current efforts involve exposing knowledge graph paths that are calculated by systems, such as KagNet~\cite{lin_kagnet_2019}, that leverage such a structure in their pipelines. We expect that this kind of information will help the systems' developers to further analyze their implementations across benchmarks.

We want to support additional use-cases. As an example, to help systems' developers with insights into the tasks that can be used to decide with which datasets may exist that may make sense for them to use for training. Similarly for tasks' creators, we want to allow them to understand their own tasks and how to augment them in order to increase coverage along certain dimensions. In addition, we want to provide them with information about dimensions not currently sufficiently evaluated, potentially indicating the need for new tasks.

We are making the MCS Benchmark ontology open and we hope that it will be adopted and potentially extended by the MCS community. It can be used to describe, compare, and explore diverse aspects of benchmarks. The MCS Benchmark ontology powers the Benchmark tool, which is a step towards a working platform for integrating diverse commonsense benchmarks, supporting a streamlined process of incorporating new or specialized tasks that challenge question-answering systems and models. It acts as a foundation for enabling implementations to access the tasks' data in a standardized way. It enables the description of benchmarks in a common vocabulary, allowing users to analyze their content, comparing these with consistent metadata across benchmarks. This ontology is in active development and available at \url{https://github.com/tetherless-world/mcs-ontology}.

\section*{Acknowledgments}

This work is funded through the DARPA MCS program award number N660011924033 to RPI under USC-ISI West.

\fontsize{9.0pt}{10.0pt} \selectfont
\bibliography{bib}

\end{document}